\documentclass[preprint,review,12pt]{elsarticle}

\usepackage{subfigure}
 \usepackage{graphics}
 \usepackage{graphicx}
\usepackage{color}
\usepackage{amssymb}
\usepackage{amsmath}
\usepackage{amsfonts}
\usepackage{multirow}
\usepackage{caption}
\usepackage{epstopdf}

\journal{Pattern Recognition}

\begin{document}

\begin{frontmatter}



\title{Quadratic Projection Based Feature Extraction \\
with Its Application to Biometric Recognition}


\author{Yan Yan~$^{a}$}
\author{Hanzi Wang~$^{a}$\corref{cor1}}
\cortext[cor1]{Corresponding author. Tel.: +86-592-2580063; fax: +86-592-2580063.\\
E-mail addresses: yanyan@xmu.edu.cn (Y.~Yan), hanzi.wang@xmu.edu.cn (H.~Wang),\\
chensi@xmut.edu.cn (S.~Chen), caoxiaochun@iie.ac.cn (X.~Cao), csdzhang@ comp.polyu.\\edu.hk (D.~Zhang)}
\author{Si Chen~$^{b}$}
\author{Xiaochun Cao~$^{c}$}
\author{David Zhang~$^{d}$}


\address{$^{a}$~Fujian Key Laboratory of Sensing and Computing for Smart City, School of Information Science and Engineering, Xiamen University, Fujian 361005, China\\
$^{b}$~School of Computer and Information Engineering, Xiamen
University of Technology, Fujian 361024, China\\
$^{c}$~State Key Laboratory of Information Security, Institute of Information Engineering, Chinese Academy of Sciences, Beijing 100093, China\\
$^{d}$~Biometrics Research Centre, The Hong Kong Polytechnic University, Hong Kong
}

\begin{abstract}
This paper presents a novel quadratic projection based feature extraction framework, where a set of quadratic matrices is learned to distinguish each class from all other classes.
We formulate quadratic matrix learning (QML) as a standard semidefinite programming (SDP) problem. However, the conventional interior-point SDP solvers do not scale well to the problem of QML for high-dimensional data. To solve the scalability of QML, we develop an efficient algorithm, termed DualQML, based on the Lagrange duality theory, to extract nonlinear features. To evaluate the feasibility and effectiveness of the proposed framework, we conduct extensive experiments on biometric recognition.
Experimental results on  three representative biometric recognition tasks, including face, palmprint, and ear recognition, demonstrate the superiority of the DualQML-based feature extraction algorithm compared to the current state-of-the-art algorithms.
\\
\end{abstract}

\begin{keyword}
Biometric recognition \sep Feature extraction \sep Quadratic projection \sep Semidefinite programming \sep Lagrange duality


\end{keyword}

\end{frontmatter}

\section{Introduction}
A typical statistical pattern recognition system usually consists of four modules: a sensor module, a preprocessing module, a feature extraction module, and a classification module \citep{Jain2004}. Among these four modules, the feature extraction module plays a critical role in the success of the system. The objective of feature extraction is to find a specific representation which encodes relevant information from input data, so that not only is the computational complexity of subsequent classifiers reduced but also the useful features can be used to perform the desired tasks \citep{Jiang2011}.

Usually, the real-world data can be represented as a high-dimensional vector \citep{Jain2007}. For instance, an image of size $80 \times 80$ can be viewed as a point in a $6,400$ dimensional feature space. However, the high dimensionality of data prevents from direct usage of learning techniques in a high-dimensional space. A common way to deal with this problem is to make use of feature extraction techniques,
or more specifically, use dimensionality reduction techniques \citep{Jiang2011,Yan2007,Harandi2014} to project the original high-dimensional data onto a low-dimensional space.

Recently, biometric recognition, which refers to the task of automatic identification of individuals based on their physiological and/or behavioral characteristics, has received much attention due to its wide range of applications, such as law enforcement, access control, and video surveillance \citep{Jain2004,Jain2007,Unar2014}. A number of biometrics have been proposed in recent years  
(e.g., \citep{Zhang2003,Yang2005,Yan2005,Xu2007, Jing2012}).
Two kinds of biometric characteristics are usually used, i.e., physiological characteristics (such as face, palmprint, and ear) and behavioral characteristics (such as gait, signature). Despite decade-long efforts, building an automatic and robust biometric recognition system remains a challenging problem due to variations in illumination, pose, occlusion, etc.

During the fast few decades, numerous feature extraction methods have been put forward to deal with the biometric recognition problems. For example, Qian et al.~\citep{Qian2013} proposed the discriminative histograms of local dominant orientation (D-HLDO) method for biometric image feature extraction. Shekhar et al.~\citep{Shekhar2014} developed a joint sparse representation for robust multimodal biometrics recognition. Beside feature extraction, feature selection is also extensively investigated to discover the knowledge related to biometric data. Different from feature extraction, which generates new features from functions of the original features, feature selection returns a subset of the features from a large feature pool.
Boosting \citep{Viola2004, Xiao2006} and Lasso \citep{Destrero2012} have been successfully used to perform feature selection in face detection and recognition. Sun et al.~\citep{Sun2014} proposed an optimization formulation for ordinal feature selection for iris and palmprint recognition.  Guo et al.~\citep{Guo2012} presented the feature band selection for the online multispectral palmprint recognition. Ghoualmi et al.~\citep{Ghoualmi2015} proposed a feature selection method based on the genetic algorithm for ear authentication. Kumar et al.~\citep{Kumar2005} suggested to use feature selection and combination to improve the performance of bimodal biometric system.

Until now, a large number of feature extraction methods have been developed. However, many methods mainly consider the first order statistics of data, which are indeed non-linearly distributed. Even though nonlinear feature extraction methods are introduced to handle the non-linearly distributions, the computational cost of these methods is high. On the other hand, high order statistics which capture the complex statistical relationship of the data
can be beneficial for feature extraction and feature selection,
potentially leading to superior performance.

In this paper, we propose a novel nonlinear feature extraction framework, which takes advantage of the quadratic projection technique. Compared with the traditional linear projection technique, the quadratic projection technique exploits the second order statistics of data. It is well-known that the quadratic classifiers are optimal for the data under Gaussian distributions. Even when the data is not Gaussian-distributed, we can still expect quadratic projection to perform better than linear projection under general conditions since more high-order information is taken into consideration in quadratic projection.

More specifically, we propose a novel nonlinear feature extraction framework based on the quadratic projection technique. Different from the traditional linear projection technique (which obtains a feature vector based on a linear form), 
the quadratic projection technique uses a quadratic form 
to extract a feature vector, where each feature is extracted by using the homogeneous polynomial of degree two in a number of original features. 
In the proposed framework, a set of quadratic matrices is learned to distinguish each class from all other classes. Mathmatically, we formulate quadratic matrix learning (QML) as a standard semidefinite programming (SDP) problem. To solve the scalability of QML, we further develop an efficient algorithm which significantly reduces the computational complexity of the conventional interior-point SDP solvers \citep{Boyd2004}.

In this paper, we will motivate and study this new framework within the context of biometric recognition. We use biometrics data as a case study to illustrate the effectiveness of the proposed framework.
 Experimental results on three representative biometric recognition tasks (including face, palmprint, and ear recognition) show that the proposed algorithm achieves better performance than the linear projection based and kernel/tensor based feature extraction algorithms.

In summary, the main contributions of our work are summarized as follows:
\begin{enumerate}
        \item
         A novel feature extraction framework based on the quadratic projection technique is proposed to extract discriminative features, where a set of quadratic matrices is learned.
         Experimental results on biometric recognition tasks show the effectiveness of the proposed framework.

        \item
        We develop an efficient algorithm for quadratic matrix learning (QML) via the Lagrange duality theory. Our proposed algorithm is much more scalable than the traditional SDP solvers. The importance of this improvement is that it thereby allows us to apply QML to high-dimensional data.

\end{enumerate}

The rest of this paper is organized as follows. Section 2 describes related work. Section 3 presents the details of the proposed quadratic projection based feature extraction framework, where a novel algorithm is developed for efficient QML. Experimental results on three biometric recognition tasks are given in Section 4. Finally, Section 5 provides the concluding remarks.
\section{Related Work}

Feature extraction can be performed in a linear or nonlinear way. The linear feature extraction based algorithms usually perform a linear mapping of input data onto a low-dimensional feature space.
Typical algorithms include principal component analysis (PCA) \citep{Turk1991}, linear discriminant analysis (LDA) \citep{Belhumeur1997,Yang2005}, locality preserving projections (LPP) \citep{He2005}, margin Fisher analysis (MFA) \citep{Yan2007}, class-dependence feature analysis (CFA) \citep{Kumar2006,Yan2008}, local discriminative Gaussians (LDG) \citep{Parrish2012}, and low rank matrix factorization \citep{Kim2015}. Recently, a large number of distance metric learning algorithms \citep{Davis2007,Weinberger2009, Wang2014,Huang2015} have been proposed to perform linear feature extraction.
These algorithms are computationally efficient. However, their performance can degrade in cases with non-linearly distributed data existing in many real-world applications.

Nonlinear feature extraction algorithms are based on the intuition that input data lies on a nonlinear manifold in a high-dimensional space. A direct and natural way to extend the linear feature extraction algorithms to nonlinear cases is to take advantage of the kernel technique \citep{Muller2001,Zafeiriou2012}, which does not have to explicitly compute the nonlinear mapping between the input space and the feature space. The kernel-based nonlinear algorithms find nonlinear projections by nonlinearly mapping data onto a higher-dimensional feature space, but it still performs linear projections in the new feature space.

Other types of nonlinear algorithms include manifold learning techniques, such as ISOMAP \citep{Tenenbaum2000}, locally linear embedding (LLE) \citep{Roweis2001}, and local tangent space alignment (LTSA) \citep{Zhang2005}. Nevertheless, many manifold learning algorithms suffer from the so-called out-of-sample problem \citep{Ham2004}, i.e., these algorithms provide mapping only for training data but not for unseen test data.

The multilinear subspace learning (MSL) techniques \citep{Lu2011,Tao2007} have also been developed for finding a low-dimensional representation of high-dimensional tensor data through direct mapping. There are three types of multilinear projections according to the forms of input and output of a projection \citep{Lu2011}, i.e., vector-to-vector projection (VVP), tensor-to-tensor projection (TTP), and tensor-to-vector projection (TVP). Although the MSL algorithms preserve the structure in original data by operating on natural tensor representations, most of these algorithms are based on iterative schemes and usually converge to local solutions.

Generally speaking, the distributions of real-world data (such as biometric data) show highly non-linear and non-convex. Therefore, the non-linear feature extraction
is beneficial for the subsequent classification. However, the kernel extension is computationally expensive, while the multi-linear based algorithms often offer local optimal solutions. In this paper, we develop a novel nonlinear feature extraction framework, which leverages the quadratic projection technique to encode high order statistics of the biometric data. Note that both the proposed algorithm and the MSL algorithms (using 2D matrix data) \citep{Lu2011,Tao2007} aim to optimize a matrix.
However, in the proposed framework, the optimized matrix is a quadratic matrix required to satisfy the positive semidefinite constraint (usually not required in the MSL algorithms). Furthermore, compared with the MSL algorithms which attempt to obtain one matrix to distinguish all classes, the proposed framework obtains multiple quadratic matrices, where each quadratic matrix is trained to separate one class from the other classes.

\section{Quadratic Projection Based Feature Extraction}
In this section, we present a quadratic projection based feature extraction framework.
We begin with an overview of the proposed framework in Section 3.1. The optimization problem of Quadratic Matrix Learning (QML) is formulated in Section 3.2. An efficient algorithm, termed DualQML, to solve the problem of QML is derived in Section 3.3. We give the complete algorithm in Section 3.4. Finally, we discuss some important issues about the proposed algorithm in Section 3.5.

Before formally presenting the proposed algorithm, we describe some notations used in this paper.
A column vector is represented by a bold lower case letter and a matrix is represented by a bold upper-case letter.
For a positive semidefinite (p.s.d.) matrix $\textbf{M}$, we denote it as $\textbf{M} \succeq 0$. Given a symmetric matrix $\textbf{A}$ and its eigen-decomposition $\textbf{A} = \textbf{U}\Sigma\textbf{U}^{\rm{T}}$, where $\textbf{U}$ is an orthonormal matrix and $\Sigma$ is a diagonal matrix, we define the positive part of $\textbf{A}$ as \citep{Boyd2004}:
\begin{align}%
\textbf{A}_{+} = \textbf{U}[\rm{max}(\bold{\Sigma}, \mathbb{O})]\textbf{U}^{\rm{T}}, \nonumber
\end{align}
and the negative part of $\textbf{A}$ as:
\begin{align}%
\textbf{A}_{-}  = \textbf{U}[\rm{min}(\bold{\Sigma},\mathbb{O})]\textbf{U}^{\rm{T}}, \nonumber
\end{align}
where, $\mathbb{O}$ is a square matrix in which all elements are zeros.  $\rm{max}(\bold{\Sigma}, \mathbb{O})$
  and $\rm{min}(\bold{\Sigma}, \mathbb{O})$ compute the element-wise maximum and minimum of two matrices, respectively. Therefore,
  the positive (or negative) part of \textbf{A} is composed of the positive (or negative) eigenvalues and the associate eigenvectors.
  Obviously, $\textbf{A} = \textbf{A}_{+} + \textbf{A}_{-}$ holds.
\subsection{Overview of the Proposed Framework}
Traditional linear feature extraction algorithms project high-dimensional
data onto a lower-dimensional feature space by using a linear projection
matrix, which computes the first order statistics of data. However,
in many real-world applications, higher order statistics of data are more
beneficial for feature extraction. In this subsection, we
propose a quadratic projection based feature extraction framework, which
exploits the homogeneous quadratic polynomials in the variables for feature extraction.

Inspired by CFA \citep{Kumar2006}, where a correlation filter is designed for each class, we propose a feature extraction framework where a quadratic matrix is learned for each class. The proposed framework contains two main steps.
First, a set of quadratic matrices is obtained, where each quadratic matrix is learned to separate a specific class from all other classes during the training stage. Then, all the learned quadratic matrices are used to perform feature extraction. More specifically, each component of a feature vector is generated by applying a quadratic projection (defined as the form of $\textbf{x}^{\mathrm{T}}\textbf{P}\textbf{x}$) to an input sample image $\textbf{x}$ ($\textbf{x}\in \Re^{m}$) according to a specific quadratic matrix $\textbf{P}$ ($\textbf{P}\in \Re^{m\times m}$ is a symmetric matrix).

As we can see, the key step of the proposed feature extraction framework is QML by which a quadratic matrix can be learned.
In the following subsections, we will describe the problem of QML in detail.
\subsection{Quadratic Matrix Learning}
Suppose that we have a set of sample images $\mathcal{S}=\{\textbf{x}_{i}\}_{i=1}^{n} \in \Re^{m}$, and given a class $c$, the sample images can be classified as:
\begin{eqnarray}
    \mathcal{I}_{c} &=& \{\textbf{x}_{i}~|~\mathrm{sample~image~}\textbf{x}_{i}~\mathrm{belonging~to~the~}c\mathrm{\text{-}th~class}\}, \nonumber \\
    \mathcal{E}_{c} &=& \{\textbf{x}_{i}~|~\mathrm{sample~image~}\textbf{x}_{i}~\mathrm{not~belonging~to~the~}c\mathrm{\text{-}th~class}\}. \nonumber
\end{eqnarray}
where $\mathcal{I}_{c}$ is the image set consisting of the intra-class sample images of the $c$-th class, while $\mathcal{E}_{c}$ is the image set consisting of the extra-class sample images of the $c$-th class.

Let us write the quadratic matrix learned for the $c$-th class as $\textbf{P}_{c}$. The objective of QML is to find a matrix so that the projected values of the data belonging to the $c$-th class and the other classes are well-separated after quadratic projections. Therefore, a simple way to define a criterion for QML is to require that the quadratic projections of the samples in $\mathcal{E}_{c}$ are minimized while at the same time, the quadratic projections of the samples in $\mathcal{I}_{c}$ should be as large as possible. This yields the following optimization criterion:
\begin{align}%
\label{EQ:1}
\min_{\textbf{P}_{c}} &~\sum_{\textbf{x}_{i}\in\mathcal{E}_{c}} \textbf{x}_{i}^{\mathrm{T}}\textbf{P}_{c}\textbf{x}_{i} \nonumber \\ \mathrm{s.~t.} &~\textbf{x}_{i}^{\mathrm{T}}\textbf{P}_{c}\textbf{x}_{i} \geq 1, \forall \textbf{x}_{i} \in \mathcal{I}_{c}  \nonumber \\
&~\textbf{P}_{c}\succeq 0.
\end{align}

Notice that $\textbf{P}_{c}$ is required to be a p.s.d matrix, which means the quadratic projections (constituting the components in the extracted feature vector) of samples are not less than 0. This is consistent with the correlation operation in CFA, where the correlation outputs (corresponding to the linear constraints during the design process of correlation filters) are non-negative. In fact, non-negative constraints of feature vectors are also beneficial for metric comparison \citep{Kumar2006}. Note that the choice of the constant on the right hand side of (1) is arbitrary. This is due to the fact that changing the constant 1 to any other positive constant $c$ will result in $\mathbf{P}_c$ being replaced by $c\mathbf{P}_c$.

However, one problem with (\ref{EQ:1}) is that it may not be suitable to solve real-world biometric recognition tasks, where data could be noisy and include a limited number of training samples as well.

To enhance the generalization capability and robustness of QML, we propose a new objective function by considering the regularization principle. It is well-known that regularization plays a critical role in many machine learning algorithms to prevent overfitting \citep{Vapnic1998}. Therefore, we propose a general regularization formulation of QML as follows:
\begin{align}%
\label{EQ:2}
\min_{\textbf{P}_{c}} &~\frac{1}{2}||\textbf{P}_{c}||_{F}^{2} + \lambda \sum_{\textbf{x}_{i}\in\mathcal{E}_{c}} \textbf{x}_{i}^{\mathrm{T}}\textbf{P}_{c}\textbf{x}_{i} \nonumber \\ 
\mathrm{s.t.} &~\textbf{x}_{i}^{\mathrm{T}}\textbf{P}_{c}\textbf{x}_{i} \geq 1, \forall \textbf{x}_{i} \in \mathcal{I}_{c} \nonumber \\
&~\textbf{P}_{c} \succeq 0,
\end{align}
where $||\textbf{P}_{c}||_{F} = \sqrt{\sum_{i,j=1}^{m}p_{i,j}^{2}}$ represents the Frobenius norm of $\textbf{P}_{c}$, if $ \textbf{P}_{c}=[p_{i,j}]_{m\times m}$.

There are two items in the objective function of (\ref{EQ:2}). The first item serves as a regularization term which prevents the value of any element within the matrix $\textbf{P}_{c}$ from being too large. The second item stands for the summed projected values corresponding to the extra-class samples. $\lambda$ is a regularized parameter to balance the two items. In addition, the first constraint in (\ref{EQ:2}) makes sure that each sample from the $c$-th class yields an output whose value is at least larger than 1. Thus, a discriminative quadratic matrix is learned such that the projected values corresponding to the intra-class samples and extra-class samples are well-separated.

To solve the above optimization problem, the second item and the first constraint in the objective function of (2) can be respectively rewritten as:
\begin{align}%
\label{EQ:3}
 \sum_{\textbf{x}_{i}\in\mathcal{E}_{c}} \textbf{x}_{i}^{\mathrm{T}}\textbf{P}_{c}\textbf{x}_{i} &= \sum_{\textbf{x}_{i}\in\mathcal{E}_{c}} tr(\textbf{P}_{c}\cdot\textbf{x}_{i}\textbf{x}_{i}^{\mathrm{T}}) \nonumber \\
&= tr(\textbf{P}_{c}\cdot\sum_{\textbf{x}_{i}\in\mathcal{E}_{c}}\textbf{x}_{i}\textbf{x}_{i}^{\mathrm{T}}) \nonumber \\
&= tr(\textbf{P}_{c} \cdot \textbf{O}_{c})
,
\end{align}
and
\begin{align}%
\label{EQ:4}
 \textbf{x}_{i}^{\mathrm{T}}\textbf{P}_{c}\textbf{x}_{i}
&= tr(\textbf{P}_{c}\cdot\textbf{x}_{i}\textbf{x}_{i}^{\mathrm{T}}),~\forall \textbf{x}_{i}\in\mathcal{I}_{c} \nonumber \\
&= tr(\textbf{P}_{c} \cdot \textbf{I}_{i}) ,~\forall \textbf{x}_{i}\in\mathcal{I}_{c},
\end{align}
where the product `$\cdot$' is a point-wise matrix multiplication operator, and $tr(\cdot)$ represents a trace operator that computes the sum of the diagonal elements of a matrix.
 $\textbf{O}_{c}$ and $\textbf{I}_{i}$ can be represented as $\sum_{\textbf{x}_{i}\in\mathcal{E}_{c}}\textbf{x}_{i}\textbf{x}_{i}^{\mathrm{T}} $ and $\textbf{x}_{i}\textbf{x}_{i}^{\mathrm{T}}$, respectively.

Thus, problem (2) can be rewritten as:
\begin{align}%
\label{EQ:5}
\min_{\textbf{P}_{c}} &~\frac{1}{2}||\textbf{P}_{c}||_{F}^{2} + \lambda tr(\textbf{P}_{c} \cdot \textbf{O}_{c}) \nonumber \\
\mathrm{s.t.} &~tr(\textbf{P}_{c} \cdot \textbf{I}_{i})\geq 1,~\forall \textbf{x}_{i}\in\mathcal{I}_{c}\nonumber \\
&~\textbf{P}_{c} \succeq 0,
\end{align}

Problem (\ref{EQ:5}) is a convex optimization problem, since the objective function is convex (this can be easily proved by using the second-order convexity conditions), the inequality constraints are linear, and the p.s.d.~constraint is convex. As a matter of fact, problem (\ref{EQ:5}) can be formulated as a standard SDP problem \citep{Vandenberghe1996}, using a standard trick which converts a quadratic objective function into a linear matrix inequality and a linear objective function. Hence, it can be directly solved by using the off-the-shelf SDP solvers \citep{Boyd2004}. However, the conventional interior-point SDP solvers suffer from a high computational complexity of $O(m^{6.5})$, where $m$ is the dimensionality of data, and it can only deal with the problems involving up to a few hundreds of variables \citep{Boyd2004}.
This hampers the application of the conventional SDP solvers to high-dimensional data, such as data in biometric recognition (usually involving thousands of variables).
\subsection{The DualQML Algorithm}
In this section, we propose to use the Lagrange duality theory \citep{Boyd2004} to make (\ref{EQ:5}) applicable to high-dimensional data.

We introduce a dual multiplier $\textbf{u}$ associated with the inequality constraints, and a matrix $\textbf{K}$ associated with the p.s.d.~constraint in the primal problem (5). According to the Lagrange duality theory,
a non-negative dual variable is associated with an inequality constraint in the primal problem. Therefore, the dual variable $\textbf{u}$ should satisfy the non-negative property. In addition, due to the fact that the p.s.d.~cone is self-dual, $\textbf{K}$ should be a p.s.d. matrix.
Hence, the Lagrangian of (\ref{EQ:5}) can be written as follows:
\begin{align}%
\label{EQ:8}
&L(\underbrace{\textbf{P}_{c}}_\text{primal}, \underbrace{\textbf{u},\textbf{K}}_\text{dual})~ \nonumber \\
&=\frac{1}{2}||\textbf{P}_{c}||_{F}^{2} + \lambda tr(\textbf{P}_{c} \cdot \textbf{O}_{c})  - \sum_{\textbf{x}_{i}\in\mathcal{I}_{c}}u_{i}tr(\textbf{P}_{c} \cdot \textbf{I}_{i}) + \sum_{i}u_{i} - tr(\textbf{P}_{c} \cdot\textbf{K})
\end{align}
with $\textbf{u}\succeq 0$ and $\textbf{K} \succeq 0$. Here $\textbf{u}\succeq 0$ denotes that all elements in $\textbf{u}$ are non-negative, and $u_{i}$ represents the $i$-th element of $\textbf{u}$.

The Karush-Kuhn-Tucker (KKT) conditions \citep{Boyd2004} are necessary and sufficient conditions for any pair of primal and dual optimal points of a convex problem. Any points that satisfy the KKT conditions are primal and dual optimal, and thus have zero duality gap. One of the KKT conditions is that the gradient of the Lagrangian with respect to the primal variable vanishes at the primal optimal point. Therefore, we can minimize the Lagrangian over $\textbf{P}_{c}$ by setting the first derivative of (6) with respect to $\textbf{P}_c$ to zero. Thus, we obtain
\begin{align}%
\label{EQ:9}
\textbf{P}_{c}^{*} =\textbf{K}^{*} -  \lambda\textbf{O}_{c} + \sum_{\textbf{x}_{i}\in\mathcal{I}_{c}}u_{i}^{*}\textbf{I}_{i}.
\end{align}
where $\textbf{P}_{c}^{*}$ and $(\textbf{u}_{i}^{*},\textbf{K}^{*})$ are respectively the primal and dual optimal solutions.

Therefore, (\ref{EQ:9}) is one KKT condition which enables us to recover the primal variable from the dual ones.

Based on the above expressions, the dual function which is defined as the minimum value of the Lagrangian over the primal variable can be obtained as follows:
\begin{align}
  & {g}(\textbf{u},\textbf{K}) = \inf_{\textbf{P}_{c}} {L} (\textbf{P}_{c}, \textbf{u},\textbf{K})  \notag \\
  &=\inf_{\textbf{P}_{c}}\frac{1}{2}||\textbf{P}_{c}||_{F}^{2} + \lambda tr(\textbf{P}_{c} \cdot \textbf{O}_{c})  - \sum_{\textbf{x}_{i}\in\mathcal{I}_{c}}u_{i}tr(\textbf{P}_{c} \cdot \textbf{I}_{i}) + \sum_{i}u_{i} - tr(\textbf{P}_{c} \cdot\textbf{K}) \notag \\
  &=\inf_{\textbf{P}_{c}}\frac{1}{2}||\textbf{P}_{c}||_{F}^{2} - tr(\textbf{P}_{c} \cdot (\textbf{K} -  \lambda\textbf{O}_{c} + \sum_{\textbf{x}_{i}\in\mathcal{I}_{c}}u_{i}\textbf{I}_{i}) ) + \sum_{i}u_{i} \notag \\
  &= \frac{1}{2}||\textbf{K} -  \lambda\textbf{O}_{c} + \sum_{\textbf{x}_{i}\in\mathcal{I}_{c}}u_{i}\textbf{I}_{i}||_{F}^{2} -
  ||\textbf{K} -  \lambda\textbf{O}_{c} + \sum_{\textbf{x}_{i}\in\mathcal{I}_{c}}u_{i}\textbf{I}_{i}||_{F}^{2} + \sum_{i}u_{i}  \notag \\
  &=  -\frac{1}{2}||\textbf{K} -  \lambda\textbf{O}_{c} + \sum_{\textbf{x}_{i}\in\mathcal{I}_{c}}u_{i}\textbf{I}_{i}||_{F}^{2} + \sum_{i}u_{i}.
\end{align}

Therefore, we obtain the Lagrange dual of (\ref{EQ:5}) as:
\begin{align}%
\label{EQ:10}
\max_{\textbf{K},\textbf{u}} &~-\frac{1}{2}||\textbf{K} -  \lambda\textbf{O}_{c}+\sum_{\textbf{x}_{i}\in\mathcal{I}_{c}}u_{i}\textbf{I}_{i}||_{F}^{2} + \sum_{i}u_{i} \nonumber \\
\mathrm{s.~t.} &~\textbf{K} \succeq 0, ~\textbf{u}\succeq 0.
\end{align}

The Lagrange dual problem (\ref{EQ:10}) is always convex, since the objective function to be maximized is concave and the constraints are convex. So, both the primal and dual problems are convex. On the other hand,
due to the convexity of the primal problem, and strict convexity of the Lagrangian with respect to the primal variable $\textbf{P}_c$, the primal problem is strictly feasible (i.e., there exist $\textbf{P}_c \succ 0$ which satisfies the linear inequalities in (5)). Slater's condition \citep{Boyd2004} is satisfied and thus strong duality between (\ref{EQ:5}) and (\ref{EQ:10}) holds. Therefore, the objective values of the two problems meet at optimality and we can obtain the solution of the primal problem by solving the dual problem.

Problem (\ref{EQ:10}) still has the p.s.d.~constraint and it is not obvious to see how to solve it in an efficient way other than using off-the-shelf SDP solvers. However, by taking the idea of alternating optimization technique, we can derive an efficient solution.
To be specific, we first fix $\textbf{u}$ and solve the optimization problem with respect to $\textbf{K}$. Then, we fix $\textbf{K}$ and solve the optimization problem with respect to $\textbf{u}$.

Given a fixed $\textbf{u}$, problem (\ref{EQ:10}) can be rewritten as:
\begin{align}%
\label{EQ:11}
\max_{\textbf{K}} &~-\frac{1}{2}||\textbf{K} -  \lambda \textbf{O}_{c}+\sum_{\textbf{x}_{i}\in\mathcal{I}_{c}}u_{i}\textbf{I}_{i}||_{F}^{2}  \nonumber \\
\mathrm{s.~t.} &~\textbf{K} \succeq 0.
\end{align}

The above optimization problem finds a p.s.d.~matrix so that $||\textbf{K} -  \lambda\textbf{O}_{c}+\sum_{i}u_{i}\textbf{I}_{i}||_{F}^{2}$ is minimized. This problem has a closed-form solution, which can be written as:
\begin{align}%
\label{EQ:12}
\textbf{K}^{*} =  (\lambda\textbf{O}_{c}-\sum_{\textbf{x}_{i}\in\mathcal{I}_{c}}u_{i}\textbf{I}_{i})_{+},
\end{align}
where $(\lambda\textbf{O}_{c}-\sum_{\textbf{x}_{i}\in\mathcal{I}_{c}}u_{i}\textbf{I}_{i})_{+}$ is the positive part of $(\lambda\textbf{O}_{c}-\sum_{\textbf{x}_{i}\in\mathcal{I}_{c}}u_{i}\textbf{I}_{i})$.
Thus, according to the definition of $(\cdot)_{+}$ , $\textbf{K}^{*}$ is a p.s.d.~matrix.

Since the optimal $\textbf{K}^*$ is expressed as a function with respect to $\textbf{u}$, the optimization problem (\ref{EQ:10}) can be simplified into a problem where only $\textbf{u}$ needs to be optimized. Therefore, we can simplify (\ref{EQ:10}) as:
\begin{align}%
\label{EQ:13}
\max_{\textbf{u}} &~-\frac{1}{2}||(\lambda\textbf{O}_{c}-\sum_{\textbf{x}_{i}\in\mathcal{I}_{c}}u_{i}\textbf{I}_{i})_{-}||_{F}^{2} + \sum_{i}u_{i} \nonumber \\
\mathrm{s.~t.} &~\textbf{u}\succeq 0.
\end{align}

Problem (\ref{EQ:13}) does not involve any matrix variables and it only has a simple constraint on $\textbf{u}$. Therefore, we can use the first-order Newton algorithm, such as L-BFGS-B \citep{Liu1989}, to solve the problem. To use L-BFGS-B, we only need to compute the gradient of the objective function of (\ref{EQ:13}), which is
\begin{align}%
\label{EQ:14}
g(u_{i}) = 1+tr((\lambda\textbf{O}_{c}-\sum_{\textbf{x}_{i}\in\mathcal{I}_{c}}u_{i}\textbf{I}_{i})_{-} \cdot \textbf{x}_{i}\textbf{x}_{i}^{\mathrm{T}}), \forall \textbf{x}_{i}\in\mathcal{I}_{c}.
\end{align}

Finally, once the optimal $\textbf{u}^*$ is obtained, the optimal $\textbf{K}^*$ can be calculated accordingly.

It is worth mentioning that the computational complexity of the proposed DualQML algorithm is much lower than the conventional SDP solvers during the training stage. This is because that at each iteration in the DualQML algorithm, the computation of (13), which runs the full eigen-decomposition, is only implemented once to obtain all the gradients. In our case, since the number of constraints is much smaller than the dimensionality of data, eigen-decomposition dominates the computational cost during each iteration. Hence, the overall computational complexity is only $O(t\cdot m^{3})$ with $t$ being around 30$\sim$50. Recall that the complexity of the conventional SDP solvers is about $O(m^{6.5})$.
Therefore, the computational cost of the proposed DualQML algorithm for training is significantly reduced, especially when the dimensionality of data is high.
\subsection{The Complete Algorithm}
As we mention previously, the key step of the quadratic projection based feature extraction framework is to obtain a set of quadratic matrices by solving the problem of QML.
We have shown the elements of the proposed DualQML based feature extraction algorithm in previous subsections.
In Algorithm 1, we give the detailed outline of the proposed algorithm for image classification.
\begin{table}[tbh!]
\centering
\vspace{3mm}
\noindent
\scalebox{0.85}{
\begin{tabular}
{p{348pt}}
\hline
\textbf{Algorithm 1}: The DualQML-based feature extraction algorithm for image classification\\
\hline
\vspace{0.05mm}
\textbf{Input:} The training data $\mathcal{S}=\{\textbf{x}_{i}\}_{i=1}^{n} \in \Re^{m}$ with $C$ classes, where $m$ is the dimensionality of data and a test image.\\ 
\textbf{Output:} The class label of the test image.
\vspace{2mm}
\\
\hline
\vspace{0.1mm}
\textbf{Training Stage:}\\
\emph{Step 1}: Do for $l$ = 1,$\cdots$, $C$: \\
 \quad \quad \quad \emph{1.1} Calculate $\textbf{O}_{c}$ and $\sum_{\textbf{x}_{i}\in\mathcal{I}_{c}}u_{i}\textbf{x}_{i}\textbf{x}_{i}^{\mathrm{T}}$ based on $\mathcal{I}_{c}$ and $\mathcal{E}_{c}$ of the\\
 \quad \quad \quad ~~~$c$-th class;\\
 \quad \quad \quad \emph{1.2} Calculate the gradient of the objective function in (\ref{EQ:14}), \\
 \quad \quad \quad ~~~and use L-BFGS-B to optimize (\ref{EQ:13});\\
 \quad \quad \quad \emph{1.3} Calculate $\sum_{\textbf{x}_{i}\in\mathcal{I}_{c}}u_{i}\textbf{x}_{i}\textbf{x}_{i}^{\mathrm{T}}$ according to the output of L-\\
  \quad \quad \quad ~~~ BFGS-B (i.e., $\textbf{u}$) and compute $\textbf{P}_{c}$ according to (\ref{EQ:9}).\\
\emph{Step 2}: Obtain the quadratic matrices for all the classes $\{\textbf{P}_{c}\}_{c=1}^{C}$.\\
\emph{Step 3}: Compute the feature matrix $\textbf{F} = [\textbf{f}_{1},\textbf{f}_{2},\cdots,\textbf{f}_{n}]$, where the \\
\quad \quad \quad ~$j$-th element in $\textbf{f}_{i}$ is written as $f_{ij}=\textbf{x}_{i}^{\mathrm{T}}\textbf{P}_{j}\textbf{x}_{i},j=1,2,\cdots,C$.\\
\vspace{0.01mm}
\textbf{Test Stage:}\\
\emph{Step 1}: Compute the feature vector $\textbf{p}$ of the test image, where the \\
\quad \quad \quad~ $j$-th element of $\textbf{p}$ is: $p_{j}=\textbf{p}^{\mathrm{T}}\textbf{P}_{j}\textbf{p},j=1,2,\cdots,C$.\\
\emph{Step 2}: Assign a class label to the test image by using the nearest\\
\quad \quad \quad~ neighbor classifier based on $\textbf{F}$ and  $\textbf{p}$.
\vspace{2mm}\\
\hline
\end{tabular}
}
\end{table}

\subsection{Discussions}
Next, we discuss a couple of important issues about the proposed algorithm. First, compared with the linear feature extraction algorithms (such as PCA, LDA), the computational complexity of the proposed algorithm is higher since an iteration scheme is used to obtain the quadratic matrix during the training stage. Nevertheless, the proposed algorithm allows for higher flexibility of the decision boundary due to the usage of a nonlinear feature extraction framework.
Second, QML is an asymmetric two-class problem since the number of extra-class samples is usually larger than that of intra-class samples. Methods to tackle the asymmetry problem include the cascade classification structure \citep{Viola2001}, AdaBoost-based algorithms \citep{Wang2012}, and asymmetric weighting of covariance matrices \citep{Jiang2009}. In contrast, during the formulation of QML, we minimize the sum of quadratic projections of the extra-class samples while constraining the quadratic projections of each intra-class sample to be larger than 1, which alleviates the overemphasis on extra-class samples.
Third, regularization is critical to ensure excellent generalization performance for many algorithms. For instance, an effective eigenspectrum regularization framework \citep{Jiang2008} was developed to extract discriminative features. In this paper, we use a Frobenius norm based regularization term to enhance the generalization and robustness performance of feature extraction, which can lead to scalable and simple optimization by considering the dual formulation.
 Finally, we note that both distance metric learning (DML) \citep{Davis2007,Weinberger2009} and QML attempt to learn a p.s.d.~matrix. However, their objective functions are intrinsically different: DML finds a metric for measuring similarity between samples, while QML learns a matrix for feature extraction.
\section{Experiments}
In this section, the performance of the proposed DualQML-based feature extraction algorithm is evaluated on three different biometric recognition tasks. Experimental configurations are presented in Section 4.1. Experiments on face recognition, palmprint recognition, and ear recognition are given in Sections 4.2, 4.3 and 4.4, respectively. The computational complexity of different methods is analyzed in Section 4.5. Finally, discussions between different algorithms are shown in Section 4.6.
\subsection{Experimental configurations}
Seven databases, including four face databases, two palmprint databases, and one ear database, are used for evaluation. We compare the proposed algorithm with several state-of-the-art linear feature extraction algorithms, including the LDA \citep{Belhumeur1997}, MFA \citep{Yan2007}, CFA with two correlation filters (i.e., OTF \citep{Kumar2006}, OEOTF \citep{Yan2008}) algorithms, and several nonlinear feature extraction algorithms, including the general tensor discriminant analysis (GTDA) \citep{Tao2007}, kernel LDA (K-LDA) \citep{Muller2001}, maximal linear embedding (MLE) \citep{Wang2011}, and the eigenspectrum regularization based kernel LDA (ER-KDA) algorithms \citep{Zafeiriou2012}. Besides, we also compare with the asymmetric principal component analysis (APCA) \citep{Jiang2009} and eigenfeature regularization and extraction (ERE) algorithms \citep{Jiang2008}, which address the asymmetric data distribution problem and the regularization problem, respectively.

All the images are normalized and cropped to the size of $32\times32$. A series of experiments is designed to compare the performance of all the competing methods under conditions with different numbers of training samples. Specifically,
in all the experiments, a subset (consisting of $m$ images per individual) of each database is randomly taken from the database to form the training set, while the rest of the database is used as the test set. For a fixed value of $m$, the experiments with randomly sampled subsets are implemented 30 times. We report the average error recognition rate and the standard variance of the achieved error rates obtained by each competing algorithm as the final results, where the lowest error recognition rate for each case is formatted in the bold font.
The regularization parameter $\lambda$ is tuned by using 10-fold cross-validation, where we set the value of the regularization parameter to be within [0.1, 10]. To be specific, the training set is randomly partitioned into $10$ equal sized non-overlapping subsets. Among the 10 subsets, a single subset is retained as the validation data for testing the model, while the remaining 9 are used as the training data. The cross-validation process is then repeated 10 times. Finally, the parameter with
the lowest recognition accuracy is chosen (similar to \citep{Huang2007}).


 Note that the training process of a quadratic matrix is to produce a correlation peak only for the authentic samples from the class of interest, which means that the maximal value criterion, i.e., the class index of the maximal component in the feature vector, can be used as the classification rule. Thus, the label of a test sample can be given according to
\begin{equation}
\label{EQ:12}
Label(\mathbf{p}) =   \rm{arg} \max_{\emph{i}=\emph{1},\cdots,\emph{C}}(\mathbf{p}[\emph{i}]),
\end{equation}
where $\mathbf{p}=(\mathbf{p}[1],\mathbf{p}[2],\cdots,\mathbf{p}[C])^{\rm{T}}$ is the extracted feature vector corresponding to the test sample.

The maximal value criterion, however, does not consider the features in the training set, which is beneficial for classification.
In this paper, the nearest neighbor (NN) classifier with the cosine similarity is also employed.
Therefore, for the proposed DualQML algorithm, we respectively evaluate the method with the cosine similarity for the NN classifier (denoted as DualQML) and that with the maximal value criterion for classification (denoted as DualQML (max)).
For the other competing algorithms, the NN classifier with the cosine similarity is employed except for APCA, where the Mahalanobis distance is used (APCA with the Mahalanobis distance performs better than that with other distances \citep{Jiang2009,Liu2002}).
\subsection{Experiments on Face Recognition}
In this section, we show the experimental results on face recognition. Four public face databases, including the AR database\footnote{http://www2.ece.ohio-state.edu/$\sim$aleix/ARdatabase.html}, PIE database\footnote{http://www.ri.cmu.edu/research\_project\_detail.html?project\_id=41\&menu\_id=261}, FERET database\footnote{http://www.itl.nist.gov/iad/humanid/feret/feret\_master.html}, and FRGC database\footnote{http://www.nist.gov/itl/iad/ig/frgc.cfm}, are used for evaluation.

\begin{figure*}[tbh!]
         \centering
            \includegraphics[width=12 cm,height=4cm]{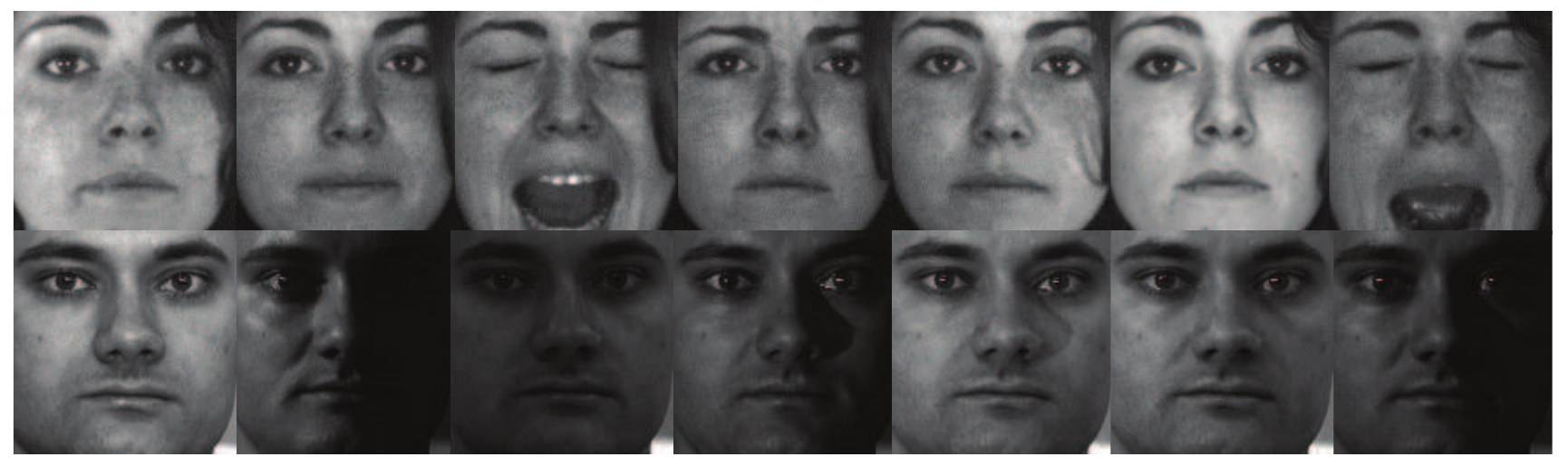}%
         \caption{Sample images of two individuals on the AR (first row) and PIE (second row) face databases.
                 }
         \label{FIG:FACE}%
        \end{figure*}

The AR database consists of over 4,000 face images from 126 individuals, including frontal views of faces with varying illumination conditions, facial expressions and occlusions. The images of most individuals were taken twice at a two-week interval. Therefore, there are two sections on the AR database, where each section contains 13 face images and 120 individuals (including 65 men and 55 women) participated on both sessions. The images of these 120 individuals are used in our experiments and only the full facial images are selected here (those facial images with occlusions are excluded since no attempt is made to handle occluded face recognition for all the competing methods). Therefore, the selected AR subset contains 120 individuals (each individual has 14 face images). The PIE database contains a large number of pose and illumination conditions along with different facial expressions. The whole PIE database has 41,368 images obtained from 68 individuals, where each individual were recorded under 43 illumination conditions, 13 poses and 3 facial expressions. Because all the competing methods mainly focus on frontal/near-frontal face recognition, we use the frontal images (with all illumination and facial expression changes) for each individual. Hence, the selected PIE subset contains 68 individuals (each individual has 46 face images).
The AR and PIE face databases are used to evaluate the performance of different methods under various illumination and facial expression changes.
Several examples on the AR and PIE face databases are shown in Fig.~\ref{FIG:FACE}.

The FERET database is a result of the FERET program sponsored by the US Department of Defense. It contains various facial expressions, illumination conditions, pose variations. To evaluate the performance of the method under small pose variations, we choose the pose subset of FERET which contains 1,400 images of 200 subjects (each subject has 7 images with pose angle ranging from $-25^{\circ}$ to $+25^{\circ}$).

\begin{figure*}[tbh!]
         \centering
            \includegraphics[width=12 cm,height=4cm]{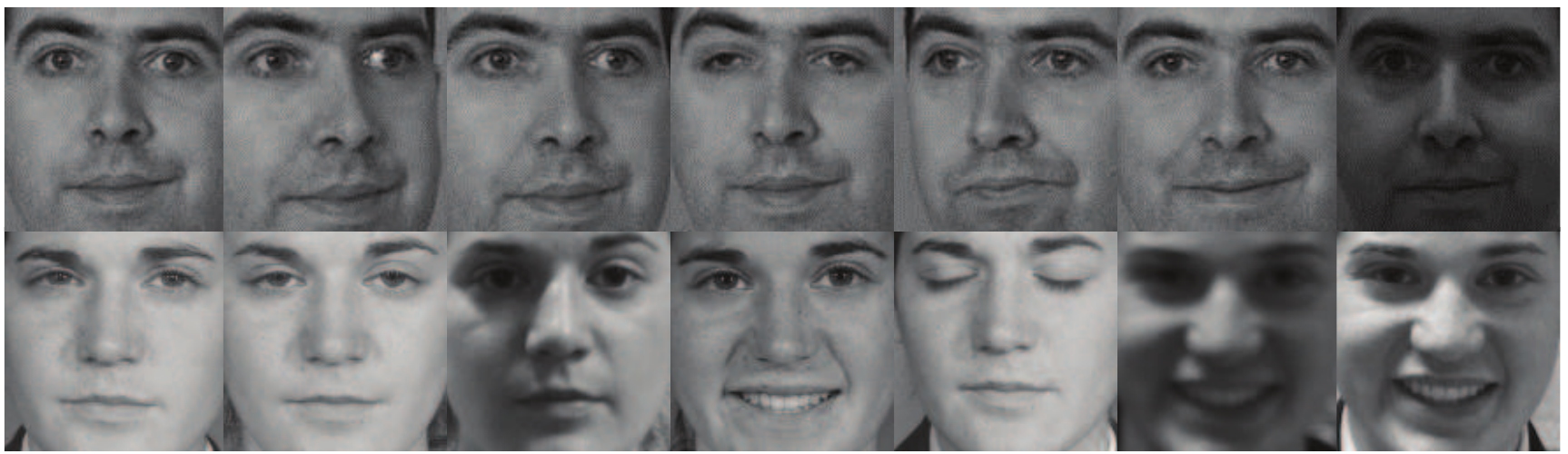}%
         \caption{Sample images of two individuals on the FERET (first row) and FRGC (second row) face databases.
                 }
         \label{FIG:FACE2}%
        \end{figure*}

The FRGC version 2.0 is a large-scale face database established under uncontrolled indoor and outdoor settings. To evaluate the performance of the method under both indoor and outdoor environments, we use 6,000 images of 300 subjects (40 images for each subject). The face images in this subset are captured in controlled and
uncontrolled conditions with severe illumination variations.
Several examples on the FERET and FRGC face databases are shown in Fig.~\ref{FIG:FACE2}. For all the databases, the values of $m$ (i.e., the number of images for each individual) to compose the training set are set to 2, 4, 6, 8, respectively (expect for FERET where the values of $m$ are set to 2, 4, 6, respectively).

        \begin{table*}[tbh!]
         \centering
         \caption
         {
          The average error recognition rates (mean\%$\pm$std.dev.) obtained by the competing algorithms on the AR and PIE face databases.
         }
         \vspace{2mm}
         \scalebox{0.65}{
         \begin{tabular}{|c|cccc|cccc|}
         \hline
         \multirow{2}{*}{Algorithm}  &\multicolumn{4}{c|}{AR} &\multicolumn{4}{c|}{PIE} \\
         \cline{2-9}  &$m=2$	 &$m=4$ 	&$m=6$	&$m=8$ &$m=2$	 &$m=4$ 	 &$m=6$	 &$m=8$ \\
         \hline
         APCA &9.35$\pm$2.2 &6.47$\pm$1.5    &5.90$\pm$2.1 &4.57$\pm$2.0 &20.11$\pm$2.1 &16.90$\pm$1.8  &13.34$\pm$1.4 &$10.54\pm$1.4\\
        \hline
          LDA  &10.88$\pm$2.1 &7.04$\pm$1.8   &6.34$\pm$1.8 &4.11$\pm$1.7
&22.54$\pm$2.3 &15.24$\pm$1.7  &13.01$\pm$1.5  &$9.56\pm$1.4\\
          \hline
         MFA  &10.86$\pm$2.2 &6.95$\pm$1.7    &6.21$\pm$2.0  &5.43$\pm$1.5
&22.98$\pm$2.4  &17.38$\pm$1.9 &14.80$\pm$1.7 &12.23$\pm$1.6\\
         \hline
          CFA-OTF  &10.02$\pm$2.0 &6.40$\pm$1.6 &5.53$\pm$1.4  &4.34$\pm$1.7
&20.01$\pm$1.7	 &16.12$\pm$1.6  &14.01$\pm$1.2 &11.86$\pm$1.5\\
         \hline
          CFA-OEOTF  &8.27 $\pm$1.8 &6.13$\pm$1.3 &4.22$\pm$1.4  &3.38$\pm$1.5
&18.43$\pm$1.4 &15.10$\pm$1.5 &$14.43\pm$1.3 &11.21$\pm$1.4
\\
          \hline
          GTDA   &10.10$\pm$2.0 &7.38 $\pm$1.4   &6.62$\pm$1.6  &.5.58$\pm$1.5
&17.11 $\pm$1.3   &15.00$\pm$1.3  &14.21$\pm$1.4 &10.76$\pm$1.5\\
         \hline
          ERE  &6.34$\pm$1.6 &4.57$\pm$\bf{0.9}  &3.86$\pm$\bf{1.1}  &2.82$\pm$\bf{1.0}
&15.24$\pm$0.8 &10.79$\pm$0.6  &8.23$\pm$0.6 &6.23$\pm$0.5\\
         \hline
          K-LDA  &12.46$\pm$2.5 &7.13$\pm$1.2 &5.65$\pm$1.5  &4.87$\pm$1.1
&21.90$\pm$1.2	 &18.10$\pm$1.1 &16.23$\pm$1.0 &14.42$\pm$1.1\\
         \hline
          MLE  &11.17$\pm$2.7 &7.10$\pm$1.2 &6.29$\pm$1.5  &5.20$\pm$1.4
&17.02 $\pm$1.1   &14.95$\pm$1.0 &12.10$\pm$1.2 &11.32$\pm$0.9\\
          \hline
          ER-KDA  &9.30$\pm$\bf{1.3} &5.26$\pm$1.1   &4.65$\pm$1.2  &3.99$\pm$1.2
&15.88 $\pm$1.2  &10.42 $\pm$1.1   &9.90$\pm$1.2 &7.54$\pm$1.0\\
          \hline
          DualQML (max) &10.80$\pm$2.1  &7.35$\pm$1.5 &5.21$\pm$1.2  &3.121$\pm$1.1
&18.24 $\pm$1.3 &15.06 $\pm$1.1 &13.00$\pm$0.9 &11.02$\pm$1.1\\
          \hline
          DualQML &\bf{6.20}$\pm$\rm{1.5}  &\bf{4.21}$\pm$\rm{1.0}  &\bf{3.39}$\pm$\rm{1.2} &\bf{1.93}$\pm$\bf{1.0}
&\bf{13.12} $\pm$0.5 &\bf{8.34}$\pm$0.6 &\bf{6.45}$\pm$0.5 &\bf{5.32}$\pm$0.4\\
          \hline
         \end{tabular}
        }
         \label{tab:face}
         \end{table*}

           \begin{table*}[tbh!]
         \centering
         \caption
         {
          The average error recognition rates (mean\%$\pm$std.dev.) obtained by the competing algorithms on the FERET and FRGC face databases.
         }
         \vspace{2mm}
         \scalebox{0.70}{
         \begin{tabular}{|c|ccc|cccc|}
         \hline
         \multirow{2}{*}{Algorithm}  &\multicolumn{3}{c|}{FERET} &\multicolumn{4}{c|}{FRGC}\\
         \cline{2-8}  &$m=2$	 &$m=4$ 	&$m=6$	&$m=2$ &$m=4$ &$m=6$ &$m=8$  \\
         \hline
         APCA & 30.25$\pm$1.5   &26.81$\pm$1.7   &20.80$\pm$1.5
&67.23$\pm$2.5 &58.90$\pm$2.4 &46.17$\pm$2.2 &35.42$\pm$2.1\\
        \hline
          LDA  &34.27$\pm$1.9   &25.18$\pm$1.6   &18.43$\pm$1.6
&65.34$\pm$2.6 &57.54$\pm$2.5    &44.75$\pm$2.4 &32.14$\pm$1.9\\
          \hline
         MFA  &31.15$\pm$2.0   &24.15$\pm$1.8   &20.64$\pm$1.9
&62.05$\pm$2.7 &56.94$\pm$2.5  &45.56$\pm$2.3 &33.11$\pm$2.2\\
         \hline
          CFA-OTF  &42.96$\pm$2.0   &27.05$\pm$1.8   &22.39$\pm$1.7
&59.43$\pm$2.2   &50.12$\pm$1.8    &42.25$\pm$1.7 &30.41$\pm$1.5\\
         \hline
          CFA-OEOTF  &28.14$\pm$1.6   &11.31$\pm$1.9   &8.12$\pm$1.7
&55.01$\pm$1.6 &45.07$\pm$1.5 &39.84$\pm$1.5 &27.91$\pm$\bf{1.0}\\
          \hline
          GTDA   &30.15$\pm$2.0   &20.72$\pm$2.2   &15.55$\pm$2.0
&68.23$\pm$2.5 &60.01$\pm$2.3 &46.96$\pm$2.4 &35.65$\pm$2.5\\
         \hline
          ERE &20.17$\pm$\bf{1.5}   &6.33$\pm$1.7   &\bf{5.19}$\pm$\rm{1.8}
&50.49$\pm$1.8 &41.94$\pm$1.7 &35.23$\pm$1.5 &25.09$\pm$1.6\\
         \hline
          K-LDA &25.33$\pm$2.1   &14.95$\pm$1.5   &10.29$\pm$1.4
&54.82$\pm$2.2 &43.93$\pm$2.0 &39.11$\pm$1.8 &26.64$\pm$1.7\\
         \hline
          MLE  &25.30$\pm$2.0   &15.13$\pm$1.3   &12.10$\pm$1.2
&58.21$\pm$2.4 &44.10$\pm$2.2 &37.95$\pm$1.9 &25.19$\pm$1.9\\
          \hline
          ER-KDA  &22.54$\pm$1.8   &10.15$\pm$1.5   &8.34$\pm$1.4
&51.13$\pm$1.7 &41.53$\pm$1.6 &34.00$\pm$1.8 &26.66$\pm$1.8\\
          \hline
          DualQML (max) &23.47 $\pm$2.1   &12.10$\pm$1.7   &11.68$\pm$1.8
&50.96$\pm$1.6 &40.21$\pm$1.4 &35.12$\pm$1.9 &25.21$\pm$1.7\\
          \hline
          DualQML &\bf{19.21} $\pm$\rm{1.6}   &\bf{6.15}$\pm$1.2   &5.43$\pm$\bf{1.0}
&\bf{43.22}$\pm$\bf{1.4} &\bf{35.14}$\pm$\bf{1.3} &\bf{29.14}$\pm$\bf{1.4} &\bf{20.45}$\pm$\rm{1.1}\\
          \hline
         \end{tabular}
        }
         \label{tab:face2}
         \end{table*}

The average error recognition rates and standard variances obtained by all the competing algorithms versus different values of $m$ on different face databases are shown in Tables \ref{tab:face} and \ref{tab:face2}. From the results, we can see that the proposed DualQML-based feature extraction algorithm achieves the best performance. Compared with DualQML (max), DualQML with the cosine similarity improves the error rates, which demonstrates the advantages of using the cosine similarity measure as a metric.
Due to the usage of the eigenspectrum regularization, ER-KDA and ERE obtain lower error recognition rates compared with the linear-based algorithms, such as LDA, MFA and CFA-OTF. In contrast, by exploiting the quadratic form, DualQML makes feature extraction more effective and discriminative.
In summary, DualQML shows more effectiveness for feature extraction in the application of face recognition than the other competing algorithms.

Note that several algorithms (such as ERE and DualQML) also achieve the good performance on the AR, PIE and FERET databases when $m=4, 6, 8$. However, these algorithms obtain higher error rates on the FRGC database which is captured under uncontrolled conditions. Therefore, how to further improve the performance of the feature extraction algorithms for databases under uncontrolled environments needs more investigation.

Both CFA (based on correlation filters)
and the proposed algorithm (based on quadratic matrices) distinguish one
specific class from all other classes for one projection axis. However, the design of
correlation filter usually uses the equality constraints while the optimization
problem of QML adopts the inequality constraints, which effectively improve
the generalization ability of the learned quadratic matrix.

\subsection{Experiments on Palmprint Recognition}
\begin{figure*}[tbh!]
         \centering
            \includegraphics[width=12 cm,height=4cm]{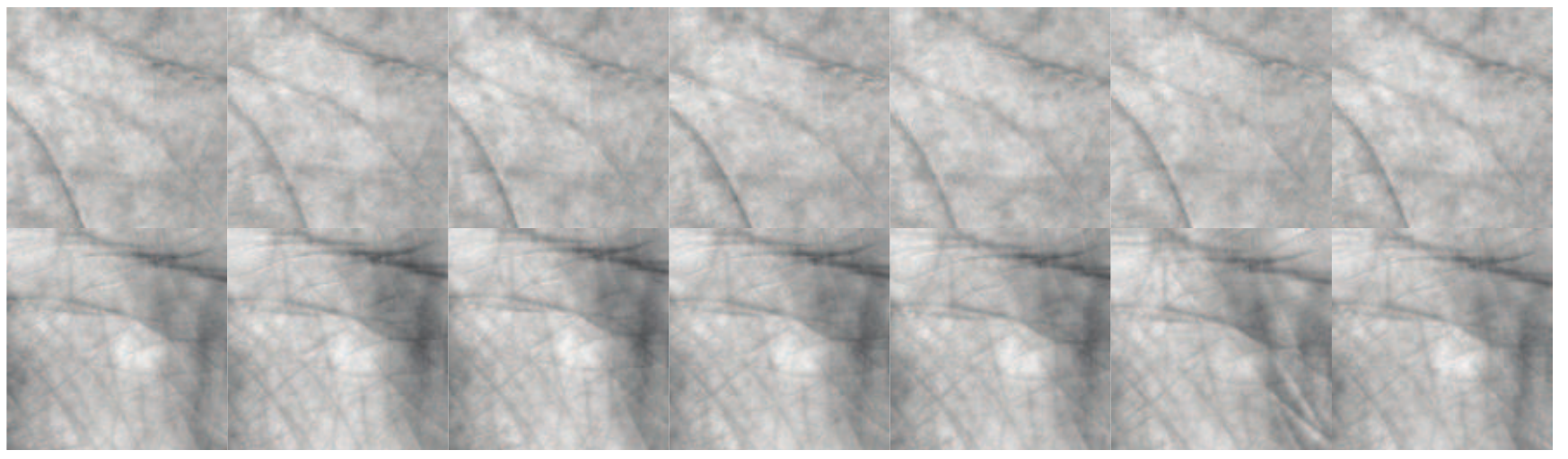}%
         \caption{Sample images of two palmprints on the PolyU (first row) and CASIA (second row) palmprint databases.
                 }
         \label{FIG:PALMPRINT}%
        \end{figure*}

In this section, we conduct experiments on palmprint recognition. The PolyU palmprint database \citep{Zhang2003} contains 7,752 gray-scale images of 386 different palms.  The CASIA palmprint database \citep{Sun2005} contains 5,502 palmprint images captured from 312 subjects. We use the two databases for evaluation. The values of $m$ to compose the training set are set to 2, 4, 6 and 8, respectively.
Several examples of the palmprint images in the database are
shown in Fig. \ref{FIG:PALMPRINT}.

        \begin{table*}[tbh!]
         \centering
         \caption
         {
          The average error recognition rates (mean\%$\pm$std.dev.) obtained by the competing algorithms on the PolyU and CASIA palmprint databases.
         }
         \vspace{2mm}
         \scalebox{0.63}{
         \begin{tabular}{|c|cccc|cccc|}
         \hline
         \multirow{2}{*}{Algorithm}  &\multicolumn{4}{c|}{PolyU} &\multicolumn{4}{c|}{CASIA}\\
         \cline{2-9}  &$m=2$	 &$m=4$ 	&$m=6$	&$m=8$  &$m=2$	 &$m=4$ 	 &$m=6$	 &$m=8$\\
         \hline
         APCA &25.61 $\pm$ 1.8 &16.19 $\pm$1.7  &9.43$\pm$1.7  &6.20 $\pm$1.5
&19.59 $\pm$2.3  &16.53$\pm$1.4  &13.29$\pm$1.2  &10.33 $\pm$1.1 \\
         \hline
          LDA &27.12$\pm$2.0    &19.50$\pm$1.8  &12.23$\pm$1.6   &9.09$\pm$1.6
&21.34 $\pm$2.2  &15.51$\pm$1.3  &12.20$\pm$1.3  &8.93 $\pm$1.2\\
          \hline
         MFA  &23.94$\pm$2.1   &15.19$\pm$1.9  &9.21$\pm$1.8   &7.40$\pm$1.7
&23.54 $\pm$1.5  &13.50$\pm$0.9  &11.49$\pm$1.2  & $8.10\pm$1.3\\
         \hline
          CFA-OTF&22.85$\pm$1.9 &15.01$\pm$1.8  &10.23$\pm$1.7   &$7.29\pm$1.6
&25.59 $\pm$2.1  &21.41$\pm$1.3  &16.23$\pm$1.4  &12.98 $\pm$1.3\\
         \hline
          CFA-OEOTF  &21.12$\pm$1.8 &13.81$\pm$1.7  &9.10$\pm$1.5   &6.93$\pm$1.5
&17.93 $\pm$2.2  &15.05$\pm$1.2  &13.10$\pm$1.1  &9.34 $\pm$1.2\\
         \hline
          GTDA   &24.15$\pm$2.1 &18.32$\pm$2.0  &13.92$\pm$1.9   &10.10$\pm$1.8
&20.54 $\pm$1.6  &17.53$\pm$1.0  &15.51$\pm$1.2  &11.29$\pm$1.0\\
         \hline
         ERE &19.27$\pm$1.6 &12.43$\pm$1.5  &9.13$\pm$1.6   &7.02$\pm$1.4
&14.56$\pm$1.3  &8.95$\pm$1.3  &6.23$\pm$1.2  &5.12$\pm$1.1\\
         \hline
          K-LDA  &26.23$\pm$2.0 &19.43$\pm$1.9  &10.90$\pm$1.6   &8.11$\pm$1.7
&19.02 $\pm$3.2  &14.95$\pm$1.1  &13.93$\pm$1.0  &10.02$\pm$0.9\\
         \hline
          MLE  &23.11$\pm$1.9  &16.53$\pm$1.8  &13.12$\pm$1.6   &11.42$\pm$1.7
& 20.64$\pm$1.8  &15.91$\pm$0.8  &13.90$\pm$1.0  &11.31$\pm$1.1\\
          \hline
          ER-KDA  &20.01$\pm$1.6 &12.96$\pm$1.7  &9.20$\pm$1.6   &7.90$\pm$1.5
&14.67 $\pm$1.5  &10.54$\pm$1.2  &$7.94\pm$1.1  &5.99$\pm$0.9\\
           \hline
          DualQML (max) &18.32$\pm$1.7 &12.10$\pm$1.5  &10.39$\pm$1.6  &8.44$\pm$1.6
&18.09 $\pm$1.8  &14.94$\pm$1.0  &12.12$\pm$1.1  &10.45 $\pm$0.8\\
          \hline
          DualQML &\bf{16.07}$\pm$1.4  &\bf{10.69}$\pm$1.3 &$\bf{7.87}\pm$\bf{1.3}  & \bf{5.13}$\pm$1.1
&\bf{13.55}$\pm$1.3  &\bf{8.78}$\pm$0.7 &\bf{5.31}$\pm$\bf{1.1}  &\bf{4.92}$\pm$0.7
\\
          \hline
         \end{tabular}
        }
         \label{tab:palm}
         \end{table*}

The experimental results are shown in Table \ref{tab:palm}. We can see that the DualQML-based feature extraction algorithm achieves the lowest error recognition rates among all the competing algorithms. In this experiment, LDA achieves high error recognition rates. This is due to the fact that the linear projection technique extracts less discrimination information than the nonlinear projection one in dealing with variations of palmprints.
Specifically, ER-KDA and ERE achieve lower error recognition rates compared with LDA, MFA, and CFA-OTF.
The performance obtained by CFA-OEOTF is better than that obtained by CFA-OTF due to the fact that CFA-OEOTF emphasizes the separation of intra-class and extra-class samples, while CFA-OTF focuses on the minimization of the correlation energy. GTDA considers an image as a tensor (i.e., a matrix), so that the internal geometric structure is kept. However, GTDA is still based on linear projections of data. Although APCA addresses the issue of asymmetric data distribution, it might not be effective to extract a compact feature set for classification. In comparison, DualQML learns different quadratic matrices for different classes. Even though the possibility of similar classes in the training set exists, the trained models of similar classes are largely different to each other. Therefore, DualQML can extract more discriminative features than the other competing algorithms.

\subsection{Experiments on Ear Recognition}
\begin{figure*}[tbh!]
         \centering
            \includegraphics[width=5 cm,height=3cm]{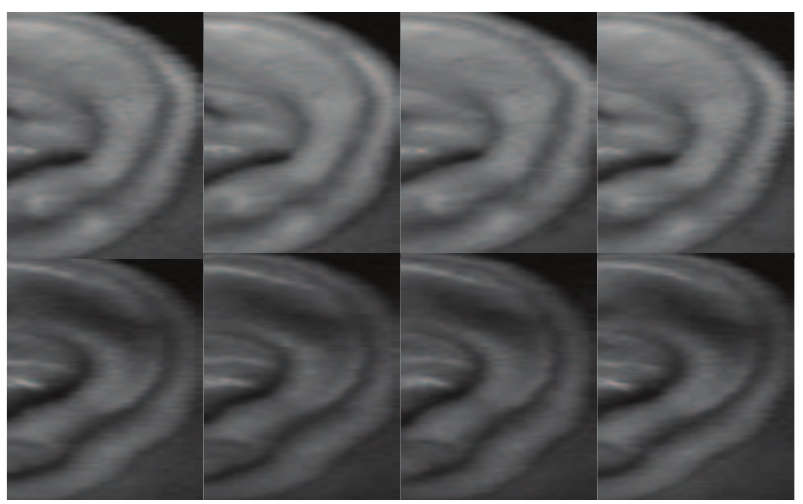}
         \caption{Sample images of two ears on the IIT Delhi ear database.
                 }
         \label{FIG:EAR}%
        \end{figure*}

In this section, we use the IIT Delhi ear database \citep{Kumar2012} for ear recognition. The database consists of the images
of 212 subjects with 754 ear images (each subject has at least three ear images). The whole database is used for evaluation.
Since some subjects in this database only have three ear images per subject, the values of $m$ to compose the training set are set to 2 and 3, respectively. Several examples of the two ear images in the database are
shown in Fig. \ref{FIG:EAR}.

       \begin{table*}[tbh!]
         \centering
         \caption
         {
          The average error recognition rates (mean\%$\pm$std.dev.) obtained by the competing algorithms on the IIT Delhi ear database.
         }
         \vspace{2mm}
         \scalebox{0.75}{
         \begin{tabular}{c|cc}
         \hline
         Algorithm &$m=2$	&$m=3$ 	\\
         \hline
         APCA &24.28$\pm$1.0 &17.90$\pm$0.9 \\
        \hline
          LDA  &24.98$\pm$0.8 &16.31$\pm$0.7 \\
          \hline
         MFA  &18.12$\pm$0.7 &15.11$\pm$0.8 \\
         \hline
          CFA-OTF  &16.88$\pm$0.7 &13.54$\pm$0.6 \\
         \hline
          CFA-OEOTF  &15.50 $\pm$0.8 &11.13$\pm$0.7 \\
          \hline
          GTDA   &16.67$\pm$0.9 &13.10 $\pm$0.8\\
          \hline
          ERE &13.98$\pm$0.9 &6.57 $\pm$0.6\\
         \hline
          K-LDA  &17.23$\pm$0.7 &14.64$\pm$0.5 \\
         \hline
          MLE  &16.69$\pm$0.8 &9.23$\pm$0.9 \\
          \hline
          ER-KDA  &16.46$\pm$0.7 &8.29$\pm$0.6\\
          \hline
          DualQML (max) &15.86$\pm$0.7  &9.13$\pm$0.9\\
          \hline
          DualQML &\bf{13.83}$\pm$0.4  &\bf{5.72}$\pm$0.4\\
          \hline
         \end{tabular}
        }
         \label{tab:ear}
         \end{table*}

The experimental results are given in Table \ref{tab:ear}. The proposed DualQML algorithm achieves the best results, with at least $\sim$2\% improvements on the error rates than all the other algorithms. Especially, APCA and LDA get the worst error recognition rates, which are much higher than the proposed DualQML. This validates that DualQML is more effective for feature extraction than APCA and LDA. The error recognition rates obtained by GTDA and K-LDA are higher than ER-KDA. This is because that ER-KDA considers the information in both the range space and the null space. ERE achieves the error rates comparable to DualQML due to an effective eigenspectrum model to alleviate problems of instability and overfitting when the number of training samples is not large.
Both MLE and DualQML are the nonlinear feature extraction methods.
However, MLE uses the combination of local linear models, which requires a large number of training samples. In contrast, DualQML considers the regularization principle to effectively handle the situation when data contain a limited number of training samples.

\subsection{Computational Complexity}
We give the computational time comparisons between the proposed
DualQML method and several representative feature extraction methods, including APCA, LDA, K-LDA, ER-KDA.
All the computational time is reported on a workstation with 2 Intel Xeon
E5620 (2.40GHz) CPUs (only one core is used) on the MATLAB platform.
Table 5 shows the total time spent on the training and the average time for testing a single image on the AR database (when $m=2$).

      \begin{table*}[tbh!]
         \centering
         \caption
         {
           Comparisons of the computational time (in seconds) used by the competing algorithms on the AR database.
         }
         \scalebox{0.8}{
         \begin{tabular}{c|c|c}
         \hline
           Algorithm &Training time	 &Average test time \\
         \hline
           APCA &65.83	&0.008	\\
          \hline
           LDA &83.54	&0.008\\
          \hline
          K-LDA &522.21	&3.501\\
          \hline
          ER-KDA &1031.53	&3.802\\
          \hline
          DualQML &5201.27	&1.431\\
         \hline
         \end{tabular}}
         \label{tab:TIME}
         \end{table*}

The computational time of DualQML
used for training is higher than that the of other methods. This is because the iterative procedure is used to obtain the quadratic matrices by considering the positive semidefinite constraint.
However, the computational time of DualQML used for test is faster than the kernel-based nonlinear algorithms, such as K-LDA, ER-KDA
(note that the time complexity of DualQML for test is $O(Cp^2+Cp)$, where $C$ is the number of classes and $p$ is the input dimensionality, while that of the kernel-based nonlinear projection based algorithms is $O(dnp)$, where $d$ is the reduced dimensionality and $n$ is the number of data points).
The proposed DualQML achieves lower error rates
compared with the competing algorithms on different biometric tasks.
On the other hand, the average test time of
the proposed algorithm is about 1 seconds per image.
As the training stage is usually performed offline,
the computational complexity of the proposed method will not limit
its applications to real-world tasks.
\subsection{Discussions}
There are two reasons to explain why the proposed DualQML algorithm shows a better performance than the state-of-the-art algorithms, such as LDA, MFA, CFA, K-LDA, MLE, and ER-KDA:

  1) The problem of DualQML is cast as a constrained optimization framework, which tries to optimize the separation between the extra-class samples and intra-class samples. LDA and MFA try to find a global projection that can maximize the between-class scatter and minimize the within-class scatter simultaneously. CFA obtains a linear projection that can discriminate one class from the other classes. Both K-LDA and ER-KDA techniques extend LDA to nonlinear projections based on the kernel technique. MLE aligns local linear models in a global coordinate space. Most methods attempt to learn a projection that shrinks distances between the same classes and expands distances between different classes in a global sense. However, the local structures in each class might not be well learned by these methods \citep{Weinberger2009}. In contrast, the proposed algorithm explicitly encourages unconstrained projected value for each sample of the class of interest, which can better adapt to different class distributions.

 2) DualQML extracts features in a class-specific manner while other algorithms extract features in a generic way. For each class in the training set, a class-specific model is learned to distinguish one class from the other classes. Based on the design criterion of QML, the features extracted from the same class are similar while those from different class are different. Therefore, DualQML can better discriminate similar classes.

\section{Conclusions}
In this paper, we have presented a novel quadratic projection based feature extraction framework and applied it to  biometric recognition. The key step is to obtain a set of quadratic matrices by solving the problem of the quadratic matrix learning (QML). To address the scalability of QML, we have developed an efficient DualQML algorithm. The key idea is that, rather than solving the primal problem, we solve the Lagrange dual problem by exploiting the special structure of QML. The proposed algorithm is simple to implement and scalable to high-dimensional biometric data. Experimental results on three types of biometric recognition tasks have shown the superiority performance of the proposed feature extraction algorithm.

\section*{Acknowledgments}
The authors would like to thank the Associate Editor and the anonymous reviewers for their constructive comments. This work was supported by the National Natural Science Foundation of China under Grants 61571379, 61472334 and 61170179 and supported by the Fundamental Research Funds
for the Central Universities under Grant 20720130720.
\end{document}